\documentclass[letterpaper, 10 pt, conference]{ieeeconf}  % Comment this line out if you need a4paper

\IEEEoverridecommandlockouts                              % This command is only needed if 
\usepackage{graphics} % for pdf, bitmapped graphics files
\graphicspath{ {./images/} }

\usepackage{epsfig} % for postscript graphics files
\usepackage{multirow}

\usepackage{amsmath} % assumes amsmath package installed
\usepackage{amssymb}  % assumes amsmath package installed
\usepackage{ulem}
\usepackage{nccmath}
 
\usepackage[utf8]{inputenc}
\usepackage[english]{babel}

\newtheorem{assumption}{\noindent\hspace{1em}\bf Assumption}

\usepackage{algorithm}
\usepackage{algpseudocode}

\usepackage{kotex}

\usepackage{xcolor}

\title{\LARGE \bf
Stable Contact Guaranteeing Motion/Force Control \\ for an Aerial Manipulator on an Arbitrarily Tilted Surface}

\author{Jeonghyun Byun$^{1}$, Byeongjun Kim$^{1}$, Changhyeon Kim$^{1}$, Donggeon David Oh$^{1}$, and H. Jin Kim$^{1}$% <-this % stops a space
\thanks{$^{1}$ The authors are with the Department of Aerospace Engineering, Seoul National University, Seoul, South Korea.
        {\tt\small \{quswjdgus97, qudwns3456,  rlackd93, donggeonoh1999, hjinkim\}@snu.ac.kr}}
}

\begin{document}

\maketitle
\thispagestyle{empty}
\pagestyle{empty}

%%%%%%%%%%%%%%%%%%%%%%%%%%%%%%%%%%%%%%%%%%%%%%%%%%%%%%%%%%%%%%%%%%%%%%%%%%%%%%%%
\begin{abstract}
This study aims to design a motion/force controller for an aerial manipulator which guarantees the tracking of time-varying motion/force trajectories as well as the stability during the transition between free and contact motions. To this end, we model the force exerted on the end-effector as the Kelvin-Voigt linear model and estimate its parameters by recursive least-squares estimator. Then, the gains of the disturbance-observer (DOB)-based motion/force controller are calculated based on the stability conditions considering both the model uncertainties in the dynamic equation and switching between the free and contact motions. To validate the proposed controller, we conducted the time-varying motion/force tracking experiments with different approach speeds and orientations of the surface. The results show that our controller enables the aerial manipulator to track the time-varying motion/force trajectories.  
\end{abstract}

%%%%%%%%%%%%%%%%%%%%%%%%%%%%%%%%%%%%%%%%%%%%%%%%%%%%%%%%%%%%%%%%%%%%%%%%%%%%%%%%
\section{INTRODUCTION}
Unmanned aerial manipulators (UAMs) interacting with structures located in hard-to-reach areas (e.g., walls or windows installed on tall structures) has been one of the most popular research topics in aerial robotics. Such tasks include window-cleaning \cite{sun2021switchable}, painting \cite{orsag2014hybrid}, and non-destructive inspection \cite{trujillo2019novel}, and there needs a motion/force controller for operations with higher precision. Particularly, for the tasks such as teleoperation \cite{allenspach2022towards}, multi-manual object manipulation \cite{shahriari2022passivity} and plug-pulling \cite{byun2021stability}, the capability to track the time-varying motion/force trajectories is also required. However, very few studies have designed a time-varying motion/force tracking controller which simultaneously considers model uncertainty and switching between the free and contact motion. 

In \cite{nava2019direct}, \cite{bodie2020active} and \cite{peric2021direct}, motion/force controllers for the omnidirectional aerial vehicles equipped with a robotic arm were proposed. However, since we are more interested in utilizing a conventional underactuated multirotor rather than developing a special configuration for an omnidirectional aerial vehicle, we focus on designing a motion/force controller for an underactuated UAM (uUAM) configured with an underactuated multirotor and a robotic arm.

An impedance-based force controller was presented in \cite{car2018impedance} and \cite{markovic2021adaptive} for a uUAM conducting peg-in-hole insertion tasks. In \cite{xu2022image}, an image-based visual impedance force controller was introduced. However, those controllers were designed under the assumption that the desired force is constant. Also, \cite{ikeda2018stable} proposed a contact force tracking controller minimizing the battery consumption, and \cite{tzoumanikas2020aerial} introduced a nonlinear motion/force model predictive controller. However, they only conducted tracking experiments for constant force. Although the tracking of time-varying force was conducted in \cite{wopereis2017application}, \cite{izaguirre2018contact} and \cite{yi2021contact}, they did not prove the stability under the model uncertainty and switching behavior between the free and contact motion.

\begin{figure}[t]
\centering
\includegraphics[width = 0.48\textwidth]{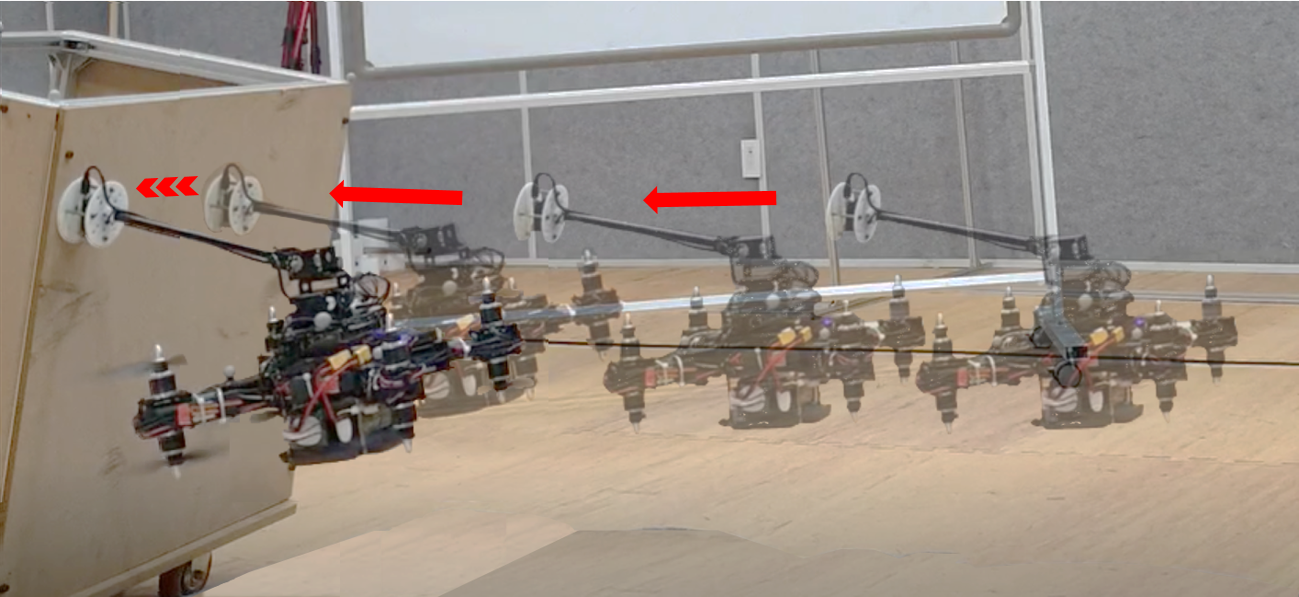}
\caption{An aerial manipulator approaches a tilted surface and tracks the desired motion and force trajectories while maintaining stable contact.} \label{fig: force exerting aerial manipulator}
\vspace{-0.15cm}
\end{figure}

In this paper, we present a motion/force controller for a uUAM which guarantees both the tracking of time-varying motion/force trajectories and stability during the transition between the free and contact motions. To this end, we derive the translational dynamic model of the uUAM exerting force on a tilted surface with respect to the position of the end-effector, and model the force as the Kelvin-Voigt linear model. Also, a disturbance-observer (DOB)-based motion/force controller is designed, and its gains are calculated to satisfy the analytically obtained input-to-state stability conditions, considering the model uncertainty as well as the switching between the free and contact motions. To validate the proposed controller, we conduct time-varying force-tracking experiments on a tilted surface with a coaxial octocopter-based aerial manipulator with different approach speeds as shown in Fig. \ref{fig: force exerting aerial manipulator}.

This paper is outlined as follows; In Section II, we formulate the translational dynamic model of a uUAM exerting force to a tilted surface, and we describe the planning and control strategies for motion/force tracking in Section III. In Section IV, we present a scheduling procedure of the controller gains which are utilized during the contact motion, and the proposed controller is validated experimentally in Section V.

\textbf{Notations:} \textit{0}$_{ij}$, $I_{i}$ and $e_3$ represent the $i\times j$ zero matrix, $i \times i$ identity matrix and $[0;0;1]$, respectively. For vectors $\alpha$ and $\beta$, we let $\alpha_i$ and $[\alpha]_{\times} \in \mathbb{R}^{3\times3}$ denote the $i$-th element of $\alpha$ and the operator which maps $\alpha$ into a skew-symmetric matrix such as $[\alpha]_{\times}\beta = \alpha \times \beta$, respectively. Also, we abbreviate the phrase "with respect to" to w.r.t..

\section{Translational Dynamic Model}
\subsection{Dynamic Equation w.r.t. the Position of the End-Effector}

\begin{figure}[t]
\centering
\vspace{0.25cm}
\includegraphics[width = 0.40\textwidth]{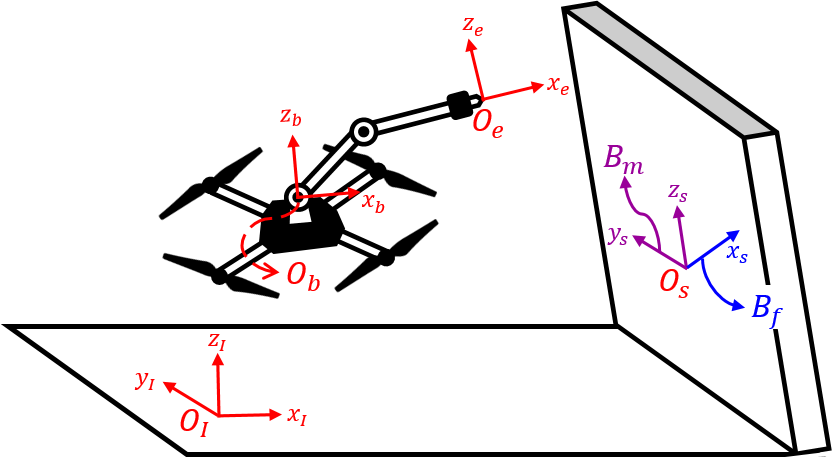}
\vspace{-0.25cm}
\caption{Illustration of a uUAM flying in front of a tilted surface. Coordinate frames required to demonstrate the uUAM system and the basis vectors of force and motion spaces are depicted.} \label{fig: aerial_manipulator_tilted_surface_frames}
\end{figure}

In Fig. \ref{fig: aerial_manipulator_tilted_surface_frames}, coordinate frames to describe the configuration of the uUAM and the tilted surface are defined. Let $O_I$, $O_b$ and $O_e$ denote the inertial, multirotor body and end-effector coordinate frames, respectively, and $O_s$ the surface coordinate frame with its $x$ axis aligned with the force exerting direction. The generalized coordinate variables of the uUAM $q \triangleq [p;\phi;\theta]$ is defined with the position of the multirotor in $O_I$, $p \triangleq [p_x;p_y;p_z]$, the Euler angles of the multirotor, $\phi \triangleq [\phi_x;\phi_y;\phi_z]$, and the joint angles of the robotic arm, $\theta \triangleq [\theta_1;\cdots; \theta_n]$. We set the generalized control input $\tau$ as $[TR(\phi)e_3 ; \tau_{\phi} ; \tau_{\theta}]$ where $T \in \mathbb{R}$, $R(\phi) \in SO(3)$ $\tau_{\phi} \in \mathbb{R}^{3}$ and $\tau_{\theta} \in \mathbb{R}^{n_a}$ represent the total thrust of the multirotor, the rotation matrix from $O_b$ to $O_I$, the body torque in the body frame and the actuator torques of the robotic manipulator, respectively, where $n_a$ means the number of actuators used in the robotic arm. 

According to \cite[Proposition 2]{yang2014dynamics} the translational dynamic model w.r.t. the center-of-mass position of a uUAM, $p_t$, is expressed as follows:
\begin{equation} \label{eqn: (citation version) translational dynamic equation}
    m_t\ddot{p}_{t} + m_tge_3 = TRe_3 + f_e + \delta_t
\end{equation}
where $m_t, g \in \mathbb{R}$, $f_e$ and $\delta_t \in \mathbb{R}^3$ express the mass of the uUAM, gravitational acceleration, force exerted on the end-effector and translational part of exogenous disturbance, respectively. To arrange (\ref{eqn: (citation version) translational dynamic equation}) w.r.t. the position of the end-effector $p_e$, we obtain the relation between $p_t$ and $p_e$ as follows:
\begin{equation} \label{eqn: rearrangement of p_t}
    \begin{split}
        p_t =& \tfrac{m}{m_t}p + \tfrac{m_e}{m_t}p_e  + \overset{n_o}{\underset{i=1}{\Sigma}}\tfrac{m_i}{m_t}p_i \\
        =& \tfrac{m}{m_t}(p_e - Rp_{be}(\theta)) + \tfrac{m_e}{m_t}p_e \\
        &+ \overset{n_o}{\underset{i=1}{\Sigma}}\tfrac{m_i}{m_t}(p_e - R(p_{be}(\theta)- p_{bi}(\theta)))\\
        =& p_e - R \ \underset{\triangleq d(\theta)}{\underline{(\tfrac{m}{m_t}p_{be}(\theta) + \overset{n_o}{\underset{i=1}{\Sigma}}\tfrac{m_i}{m_t}(p_{be}(\theta)-p_{bi}(\theta)))}}  
    \end{split}
\end{equation}
where $m$, $m_e$ and $m_i$ mean the mass of the multirotor, the end-effector and the $i$-th object of the robotic manipulator, respectively, and $n_o$ represents the number of the added objects. Also, we let $p_{be}(\theta)$ and $p_{bi}(\theta)$ denote the positions of the end-effector and the $i$-th object w.r.t. $O_b$, respectively. If we differentiate $p_t$ twice w.r.t. time and substitute it for (\ref{eqn: (citation version) translational dynamic equation}), the translational dynamic equation w.r.t. $p_e$ is formulated as follows:
\begin{equation} \label{eqn: (p_e version) translational dynamic equation}
    m_t\ddot{p}_e = -m_tge_3 + u_e + f_e + \Delta_e
\end{equation}
where
\begin{equation*}
    \begin{split}
        \Delta_e \triangleq& m_tR(([\omega_b]_{\times}^2+[\omega_b]_{\times})d+2[\omega_b]_{\times}\dot{d}+\ddot{d}) + \delta_t, \\
        u_e \triangleq& TRe_3 = T\Psi\Phi, \\
        \Psi \triangleq& \begin{bmatrix}c_{\phi_z} & s_{\phi_z} & 0 \\
        s_{\phi_z} & -c_{\phi_z} & 0 \\ 
        0 & 0 & 1 \end{bmatrix}, \quad \Phi \triangleq \begin{bmatrix} c_{\phi_x}s_{\phi_y} \\ s_{\phi_x} \\ c_{\phi_x}c_{\phi_y}\end{bmatrix}.
    \end{split}
\end{equation*}
with the angular velocity of $O_b$ w.r.t. $O_I$ expressed in $O_b$, $\omega_b$.

\subsection{Control Input Extraction}
In (\ref{eqn: (p_e version) translational dynamic equation}), $u_e$ acts as a control input. However, since the roll and pitch angles cannot be manually set, there needs the following assumption on the relation between $(\phi_x(t), \phi_y(t))$ and its reference trajectory $(\phi_{x,r}(t), \phi_{y,r}(t))$ as follows:
\begin{assumption} \label{asm: assumption on reference attitude trajectories}
    Attitude controller is properly designed so that the roll and pitch angles $\phi_x(t)$ and $\phi_y(t)$ follow their reference trajectories $\phi_{x,r}(t)$ and $\phi_{y,r}(t)$ as follows: 
    \begin{equation}
        \begin{split}
            \phi_{x}(t) = \phi_{x,r}(t-\gamma_{\phi_x}), \quad \phi_{y}(t) = \phi_{y,r}(t-\gamma_{\phi_y})
        \end{split} 
    \end{equation}
    with time-varying nonnegative delays $\gamma_{\phi_x}$ and $\gamma_{\phi_y}$.
\end{assumption}

According to \cite{lee2019robust}, if we simply replace $(\phi_x(t), \phi_y(t))$ in $u_e$ into $(\phi_{x,r}(t), \phi_{y,r}(t))$, respectively, there might be control performance degradation due to the error in roll and pitch angles. If we let $\bar{u}_e$ denote the desired value of $u_e$ calculated by a well-designed position controller, we extract the desired roll, pitch angles and total thrust ($\bar{T}$) as follows:
\begin{equation} \label{eqn: extracting the total thrust and the reference trajectories of the roll & pitch angles}
    \begin{split}
        \bar{T} =& \tfrac{(\Psi\bar{u}_{e})_3}{c_{\phi_x}c_{\phi_y}} \\
        \phi_{x,r} =& \sin^{-1}(\tfrac{(\Psi\bar{u}_{e})_2}{\bar{T}}), \quad \phi_{y,r} = \sin^{-1}\big(\tfrac{(\Psi\bar{u}_{e})_1}{\bar{T}c_{\phi_x}}\big).
    \end{split}
\end{equation}

\subsection{Dynamic Equation Decomposition}
To conduct motion/force control, the translational dynamic model (\ref{eqn: (p_e version) translational dynamic equation}) is decomposed into two parts, force and motion, as introduced in \cite{byun2022hybrid}:
\begin{align}
    m_t\ddot{x}_f =& -m_tgB^{\top}_fe_3 + u_f + f_f + B^{\top}_f\Delta_e \label{eqn: dynamic equation of the force space} \\
    m_t\ddot{x}_m =& -m_tgB^{\top}_me_3 + u_m + f_m + B^{\top}_m\Delta_e \label{eqn: dynamic equation of the motion space}
\end{align}
where 
\begin{equation*}
    \begin{split}
        x_f \triangleq& B^{\top}_fp_e, \quad u_f \triangleq B^{\top}_fu_e, \quad f_f \triangleq B^{\top}_ff_e, \\
        x_m \triangleq& B^{\top}_mp_e, \quad u_m \triangleq B^{\top}_mu_e  \quad f_m \triangleq B^{\top}_mf_e
    \end{split}
\end{equation*}
with the basis vector of the force part,  $B_f \in \mathbb{R}^{3}$, and the concatenation of the basis vectors of the motion part,  $B_m \in \mathbb{R}^{3 \times 2}$. In (\ref{eqn: dynamic equation of the force space}), the force exerted to the end-effector that is normal to the tilted surface, $f_f$, is expressed with the following Kelvin-Voigt linear model:
\begin{equation} \label{eqn: Kelvin-Voigt linear model}
    f_f = -k_e(x_f - x_{f,s}) - b_e\dot{x}_f
\end{equation}
where $x_{f,s} \triangleq B^{\top}_fp_s$ where $k_e, b_e \in \mathbb{R}$ and $p_s \in \mathbb{R}^{3}$ represent the environment stiffness, environment damping coefficient and the position of the contact point w.r.t. $O_I$, respectively. Even though $f_f$ is measured by 1DOF force sensor, the friction force tangential to the surface, $f_m$, is treated as an exogenous disturbance. Hence, because $B_f$ and $B_m$ are orthonormal to each other as shown in Fig. \ref{fig: aerial_manipulator_tilted_surface_frames}, (\ref{eqn: dynamic equation of the force space}) and (\ref{eqn: dynamic equation of the motion space}) are rearranged as follows:
\begin{align}
    m_t\ddot{x}_f =& -m_tgB^{\top}_fe_3 + \bar{u}_f + f_f + \Delta_f \label{eqn: rearranged dynamic equation of the force space} \\
    m_t\ddot{x}_m =& -m_tgB^{\top}_me_3 + \bar{u}_m + \Delta_m \label{eqn: rearranged  dynamic equation of the motion space}
\end{align}
where
\begin{equation*}
    \begin{split}
        \bar{u}_f \triangleq& B^{\top}_f\bar{u}_e, \quad \Delta_{f} \triangleq \bar{u}_f - u_f + B^{\top}_f\Delta_e, \\
        \bar{u}_m \triangleq& B^{\top}_m\bar{u}_e, \quad \Delta_m \triangleq \bar{u}_m - u_m + f_m + B^{\top}_m\Delta_e.
    \end{split}
\end{equation*}

\section{Controller Design}
In this section, the structure of the motion/force controller is presented. To this end, we first estimate $k_e$ and $b_e$ in (\ref{eqn: Kelvin-Voigt linear model}) using recursive least-squares estimation (RLSE). Then, we generate the reference trajectories of $x_f$, $f_f$ and $x_m$ and calculate the control inputs $\bar{u}_f$ and $\bar{u}_m$.

\subsection{Environment Parameter Estimation}
As introduced in \cite{lin2021unified}, we estimate $k_e$ and $b_e$ as follows:
\begin{equation} \label{eqn: environment parameter estimation}
    \begin{split}
        \epsilon &\triangleq f_f - Y\hat{\theta}_e, \qquad \dot{\hat{\theta}}_e = P Y^{\top} \epsilon \\
        \dot{P} &= \begin{cases} 
            \mu_1 P - \mu_2 P Y^{\top} Y P, & \lambda_{\textrm{max}}(P) \leq \rho_M \\
            \textit{0}_{21}, & \textrm{otherwise}
        \end{cases}
    \end{split}
\end{equation}
where $Y \triangleq -[x_f-B^{\top}_fp_s \ \dot{x}_f]$, $\hat{\theta}_e \triangleq [\hat{k}_e ; \hat{b}_e]$ and $\lambda_{\textrm{max}}(\cdot)$ represents the maximum eigenvalue of a square matrix with a large positive number $\rho_M$. Since the undesirable peaking in $\hat{\theta}_e$ can hinder the generation of reference motion/force trajectories, we set its lower and upper bounds as $\hat{k}_e \in [k_{e,m}, k_{e,M}]$ and $\hat{b}_e \in [b_{e,m}, b_{e,M}]$.

\subsection{Reference Motion/Force Trajectories Generation}
To enhance the control performance, we generate smooth reference trajectories of motion and force $x_{f,r}(t)$, $x_{m,r}(t)$ and $f_{f,r}(t)$ from their setpoints $x_{f,d}(t)$, $x_{m,d}(t)$ and $f_{f,d}(t)$ \cite{heck2016guaranteeing}. When the uUAM is flying in the free space, the reference trajectories are generated as follows:
\begin{equation} \label{eqn: reference motion-force trajectory smoothing during the free-motion}
    \begin{split}
        \ddot{x}_{f,r} &= -2\omega_n\dot{x}_{f,r} - \omega_n^2(x_{f,r}-x_{f,d}), \quad f_{f,r} = 0,\\
        \ddot{x}_{m,r} &= -2\omega_n\dot{x}_{m,r} - \omega_n^2(x_{m,r}-x_{m,d})
    \end{split}
\end{equation}
with a natural frequency $\omega_n$. Meanwhile, when the uUAM is in the contact motion, the reference trajectories are generated as follows:
\begin{equation} \label{eqn: reference motion-force trajectory smoothing during the contact-motion}
    \begin{split}
        \ddot{x}_{f,r} &= - \tfrac{\bar{k}_e}{\bar{b}_e}\dot{x}_{f,r} - \tfrac{1}{\bar{b}_e}\dot{f}_{f,r},\\
        \ddot{f}_{f,r} &= -2\omega_n\dot{f}_{f,r} - \omega_n^2(f_{f,r}-f_{f,d}),\\
        \ddot{x}_{m,r} &= -2\omega_n\dot{x}_{m,r} - \omega_n^2(x_{m,r}-x_{m,d}).
    \end{split}
\end{equation}

\subsection{DOB-based Motion/Force Controller}
\subsubsection{Control Law}
Let $\bar{m}_t$ and $\bar{g}$ denote the nominal values of $m_t$ and $g_t$, respectively, the switching control laws for $\bar{u}_m$ and $\bar{u}_f$ are shown as follows:
\begin{equation} \label{eqn: switching control laws}
    \begin{split}
        \bar{u}_f =& \begin{cases} \bar{m}_t\ddot{x}_{f,r} + k_{d}\dot{e}_{x,f} + k_{p}e_{x,f} \\
        \quad + \bar{m}_t\bar{g}B^{\top}_{f}e_3 - \hat{\Delta}_{f}, \quad \textrm{(free motion)} \\
        \bar{m}_t\ddot{x}_{f,r} - f_{f,r} - k_fe_{f,f} + b_f\dot{e}_{x,f} \\
        \quad + \bar{m}_t\bar{g}B^{\top}_{f}e_3 - \hat{\Delta}_{f},\quad \textrm{(contact motion)}
        \end{cases} \\ 
        \bar{u}_m =& \bar{m}_t\ddot{x}_{m,r} + K_{m,d}\dot{e}_{x,m} + K_{m,p}e_{x,m} \\
        &\quad + \bar{m}_t\bar{g}B^{\top}_{m}e_3 - \hat{\Delta}_{m}
    \end{split}
\end{equation}
where $e_{x,f} \triangleq x_{f,r} - x_f$, $e_{x,m} \triangleq x_{m,r} - x_m$ and $e_{f,f} \triangleq f_{f,r} - f_f$ with the user-defined positive parameters $k_{p}$, $k_{d}$ and positive definite matrices $K_{m,p}, K_{m,d} \in \mathbb{R}^{2 \times 2}$. Also, $\hat{\Delta}_f$ and $\hat{\Delta}_m$ are the estimated disturbances from the DOBs w.r.t. the force and motion spaces, and $k_f$ and $b_f$ are the \textbf{force-controller gains} calculated from the \textbf{force-controller-gain scheduler} which will be explained in Section IV.

\subsubsection{DOB (Disturbance Observer)}
$\hat{\Delta}_{f}$ and $\hat{\Delta}_{m}$ are obtained from the DOBs formulated as follows :
\begin{equation} \label{eqn: disturbance observer}
    \begin{split}
        {\nu}_{f} &= \bar{m}_{t}L_{f}\dot{x}_{f}, \quad \hat{\Delta}_{f} = z_{f} + {\nu}_{f} \\
        \dot{z}_{f} &= -L_{f}z_{f} + L_{f}(\bar{m}_t\bar{g}B^{\top}_{f}e_3 - f_{f} \bar{u}_{f} - {\nu}_{f}) \\
        {\nu}_{m} &= \bar{m}_{t}L_{m}\dot{x}_{m}, \quad \hat{\Delta}_{m} = z_{m} + {\nu}_{m} \\
        \dot{z}_{m} &= -L_{m}z_{m} + L_{m}(\bar{m}_t\bar{g}B^{\top}_{m}e_3 - \bar{u}_{m} - {\nu}_{m})
    \end{split}
\end{equation}
where $L_f \in \mathbb{R}$ and $L_m \in \mathbb{R}^{2 \times 2}$ represent a positive parameter and a positive definite matrix, respectively. According to \cite{mohammadi2013nonlinear}, if $\dot{\Delta}_{f}$ and $\dot{\Delta}_{m}$ are bounded, $||\hat{\Delta}_{f} - \Delta_{f}||$ and $||\hat{\Delta}_{m} - \Delta_{m}||$ exponentially converge to the balls with certain radius.

\section{Force-Controller-Gain Scheduler}
In this section, we first derive the input-to-state stability conditions for the force-controller gains. Then, with given $k_p$, $k_d$, $\hat{k}_e$ and $\hat{b}_e$, we set $k_f$ and $b_f$ to the values located at the farthest point from the boundary that distinguishes when the given switched system is stable and unstable.

\subsection{Input-to-State Stable (ISS) Conditions}
By substituting (\ref{eqn: Kelvin-Voigt linear model}) and (\ref{eqn: switching control laws}) for (\ref{eqn: rearranged dynamic equation of the force space}), the perturbed switched system is obtained as follows:
\begin{equation} \label{eqn: perturbed switched system of the force space}
    \begin{split}
        \dot{z}_f = A_{i}z_f + Nw_{i}(t) = \begin{bmatrix} 0 & 1 \\ -K_{i} & -B_{i} \end{bmatrix}z_f + Nw_i(t) \\
        z_f \in \Omega_i(t), \quad i \in \{1, 2\},
    \end{split}
\end{equation}
where $z_f \triangleq [e_{x,f} ; \dot{e}_{x,f}]$, $N \triangleq [0;1]$,
\begin{equation*}
    \begin{split}
        K_1 \triangleq& \tfrac{k_{p}}{m_t}, \quad B_1 \triangleq \tfrac{k_{d}}{m_t}, \\
        K_2 \triangleq& \tfrac{(1+k_f)k_e}{m_t}, \quad B_2 \triangleq \tfrac{(1+k_f)b_e+b_f}{m_t},\\
        w_1(t) \triangleq& -\tfrac{\tilde{m}_t\ddot{x}_{f,r} +(\bar{m}_t\bar{g}-m_tg)B^{\top}_fe_3 - \tilde{\Delta}_f}{m_t}, \\
        w_2(t) \triangleq& - \tfrac{\tilde{m}_t\ddot{x}_{f,r} + (1+k_f)(\tilde{b}_e\dot{x}_{f,r} + \tilde{k}_e(x_{f,r} - x_{f,s}))}{m_t} \\
        &- \tfrac{(\bar{m}_t\bar{g}-m_tg)B^{\top}_fe_3 - \tilde{\Delta}_f}{m_t},\\
        \Omega_1 \triangleq& \{z_f \in \mathbb{R}^2 \ | \ x_{f,r} - z_{f,1} \leq 0\},\\
        \Omega_2 \triangleq& \{z_f \in \mathbb{R}^2 \ | \ 0 < x_{f,r} - z_{f,1}\}
    \end{split}
\end{equation*}
with $\tilde{m}_t \triangleq \bar{m}_t - m_t$, $\tilde{k}_e \triangleq \hat{k}_e - k_e$, $\tilde{b}_e \triangleq \hat{b}_e - b_e$ and $\tilde{\Delta}_f \triangleq \hat{\Delta}_f - \Delta_f$.

According to \cite[Theorem 1]{heck2016guaranteeing}, with $\Delta K \triangleq K_1 - K_2$ and $\Delta B \triangleq B_1 - B_2$, the solution of (\ref{eqn: perturbed switched system of the force space}), $z_f(t)$, is ISS w.r.t. $w_i(t)$ if $K_1$, $B_1$, $K_2$ and $B_2$ satisfy at least one of the following conditions:
\begin{itemize}
    \item \textbf{No switching conditions}: Stable transition from free to contact motion without detaching
    \begin{enumerate}
        \item $\Delta B < 0, 4K_1 \leq B^2_1 \ \textrm{and} \ \tfrac{\Delta K}{\Delta B} < \tfrac{2K_1}{B_1-\sqrt{B_1^2 - 4K_1}}$
        \item $\Delta B < 0$, $4K_2 \leq B^2_2$ and $\tfrac{2K_2}{B_2+\sqrt{B^2_2 - 4K_2}}<\tfrac{\Delta K}{\Delta B}$ 
        \item $0 \leq \Delta B$ and $4K_2 \leq B^2_2$ 
    \end{enumerate}
    \item \textbf{Finite switching condition}: Finite number of switches between free and contact motion before achieving the stable contact.
    \begin{itemize}
        \item $\Lambda_1 \Lambda_2 < 1$ where $\Lambda_i$, $i \in \{1, 2\}$, are defined as: \begin{enumerate}
        \item if $B_i^2 < 4K_i$, 
        \begin{equation*} \label{eqn: Lambda definition 1}
            \big(\tfrac{K_i}{\omega_i} \big(\tfrac{(\Delta K)^2}{L^2} + \tfrac{Q^2}{4\omega_i^2L^2}\big)^{-\tfrac{1}{2}})^{(-1)^i}e^{-\tfrac{-B_i}{2\omega_i}\varphi_i}
        \end{equation*} 
         with $L \triangleq \sqrt{(\Delta K)^2 + (\Delta B)^2}$, $\varphi_i \triangleq \textrm{mod}\Big(-\tan^{-1}\Big(\tfrac{(-1)^i2\omega_i\Delta K}{Q}),\pi\Big)$, $\omega_i \triangleq \tfrac{1}{2}\sqrt{4K_i-B_i^2}$ and $Q \triangleq B_i\Delta K - 2K_i \Delta B$,
        \item if $B_i^2 = 4K_i$, $|\tfrac{B_iL}{2\Delta K-B_i\Delta B}|$
        \item if $B_i^2 > 4K_i$,
        \begin{equation*} \label{eqn: Lambda definition 3}
            |\tfrac{\Delta K\lambda_{bi}+K_i\Delta B}{K_iL}|^{\tfrac{(-1)^i\lambda_{ai}}{\lambda_{bi}-\lambda_{ai}}} |\tfrac{\Delta K\lambda_{ai}+K_i\Delta B}{K_iL}|^{\tfrac{(-1)^i\lambda_{bi}}{\lambda_{ai}-\lambda_{bi}}}
        \end{equation*} 
        with $\lambda_{ai} \triangleq \tfrac{-B_i - \sqrt{B_i^2-4K_i}}{2}$ and $\lambda_{bi} \triangleq \tfrac{-B_i + \sqrt{B_i^2-4K_i}}{2}$.
    \end{enumerate}
    \end{itemize}
\end{itemize}

If the above ISS conditions are satisfied, $z_f(t)$ is bounded to a small ball around the origin because $w_1(t)$ and $w_2(t)$ are also bounded due to the smoothness of reference trajectories $x_{f,r}$, $\dot{x}_{f,r}$ and $\ddot{x}_{f,r}$.

\subsection{Force-Controller-Gain Scheduler}
Prior to calculating the force-controller gains, due to the motor saturation and the noise in velocity measurement, we need to set the limits of $k_f$ and $b_f$ as follows:
\begin{equation} \label{eqn: limits on k_f and b_f}
    0 < k_{f,m} \leq k_f \leq k_{f,M}, \quad 0 < b_{f,m} \leq b_f \leq b_{f,M}.
\end{equation}

\begin{figure}[t]
\centering
\vspace{0.2cm}
\includegraphics[width = 0.48\textwidth]{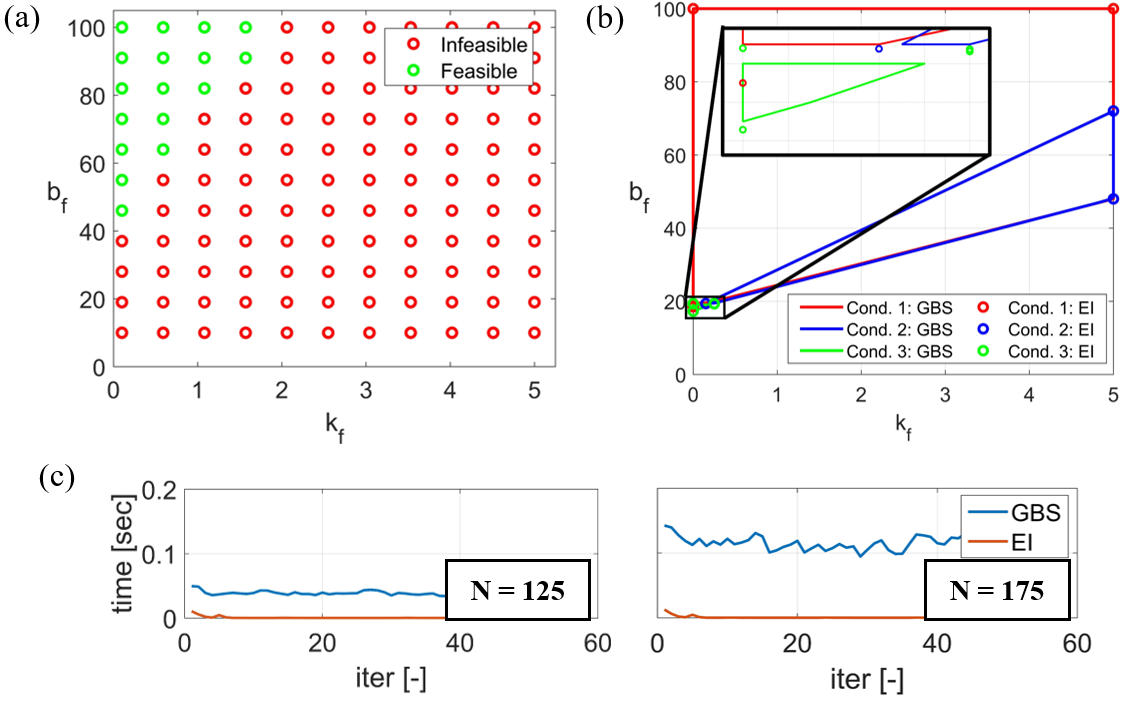}
\vspace{-0.65cm}
\caption{(a) Example of grid-based search of stable region of force-controller gains ($N = 20$), (b) Computed results of region of stable force-controller gains by the grid-based search (GBS) and the method using explicit inequalities (EI) and (c) Comparison of computation time for searching stable regions of the force-controller gains when $N = 125$ and $N = 175$.} \label{fig: GBS and comp time}
\vspace{-0.15cm}
\end{figure}

With (\ref{eqn: limits on k_f and b_f}), the procedure of force-controller-gain scheduling is summarized as follows:
\begin{enumerate}
    \item Find the convex hulls which envelop the \textit{no switching conditions} 1), 2) and 3), respectively.
    \item Find a convex hull with the largest area and set $k_f$ and $b_f$ to its center of geometry.
    \item If all three convex hulls are empty sets, find $k_f$ and $b_f$ which minimize the cost function $J(k_f, b_f)$ defined as follows (related to the \textit{finite switching condition}):
    \begin{equation} \label{eqn: cost function of the fourth ISS condition}
        \begin{split}
            J(k_f, b_f) \triangleq \Lambda_1\Lambda_2 &+ (\tfrac{2}{k_{f,M}-k_{f,m}})^2(k_f - \tfrac{k_{f,M}+k_{f,m}}{2})^2 \\
            &+ (\tfrac{2}{b_{f,M}-b_{f,m}})^2(b_f - \tfrac{b_{f,M}+b_{f,m}}{2})^2.    
        \end{split}
    \end{equation}
    \item If $k_f$ and $b_f$ do not exist, set $k_f$ and $b_f$ to $k_{f,m}$ and $k_d$, respectively.
\end{enumerate}
To proceed with step 1), we need to find the region of force-controller gains which satisfy each of the \textit{no-switching conditions}. The most straightforward way is to generate $(N+1)^2$ grid points in the rectangular area represented by (\ref{eqn: limits on k_f and b_f}) and check whether the ISS conditions are satisfied as in Fig. \ref{fig: GBS and comp time}a. However, because the time complexity of this method is $\mathcal{O}(N^2)$, the stable region of force-controller gains may not be obtained within a controller loop with large $N$. Therefore, we rearrange the \textit{no-switching conditions 1), 2) and 3)} into explicit inequalities, e.g., $f(k_f) \leq b_f$, and find the grid points that comprise the convex hull of each condition. Fig. \ref{fig: GBS and comp time}b compares the computation time of the grid-based search algorithm and the method using the explicit inequalities, where the latter is much faster. Also, Fig. \ref{fig: GBS and comp time}c shows the comparative results of those two methods. Meanwhile, to proceed with step 3), we adopt the gradient-free optimization algorithms such as \textit{pattern search} \cite{audet2002analysis}.

\begin{figure}[t]
\centering
\includegraphics[width = 0.275\textwidth]{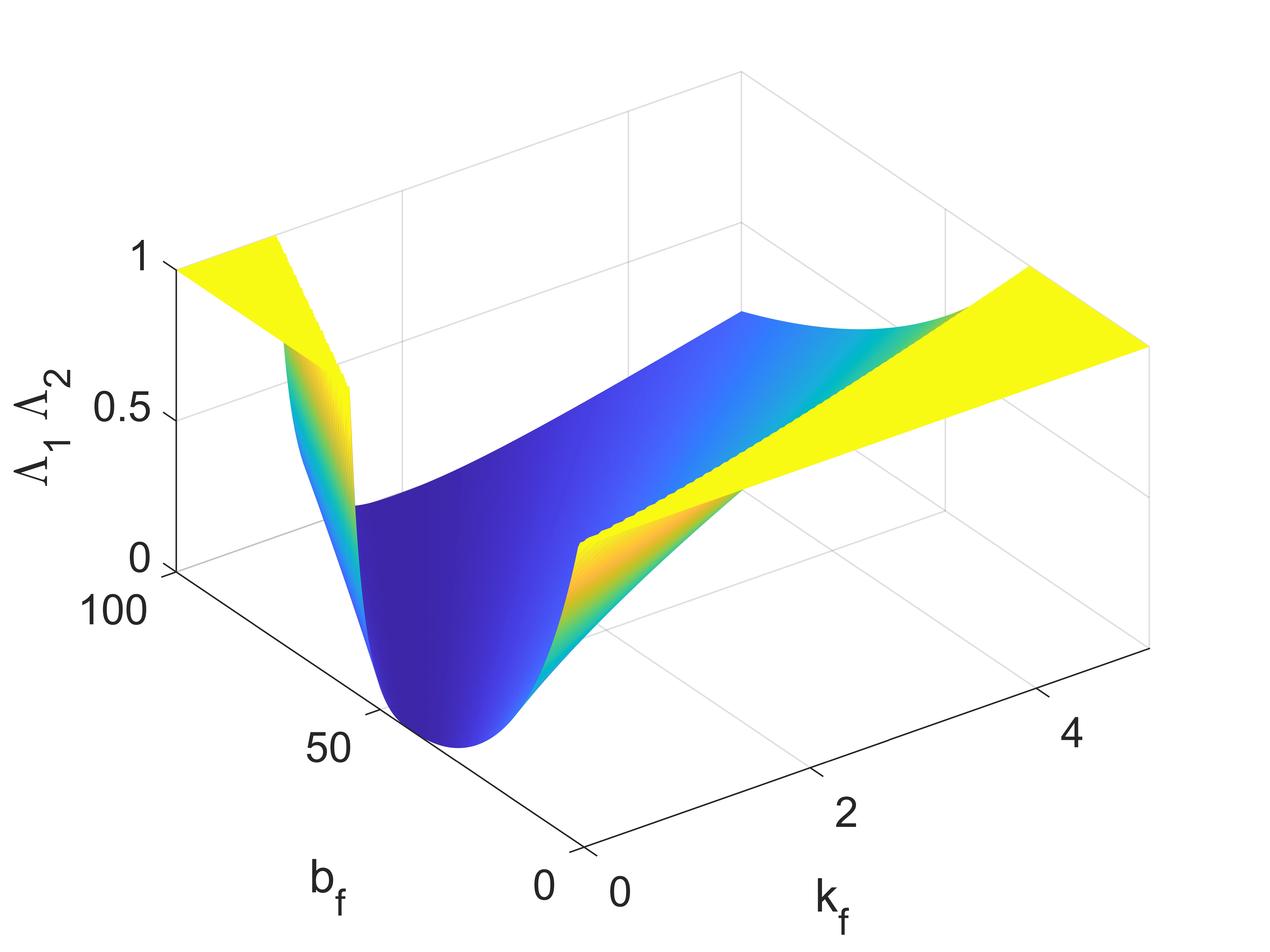}
\vspace{-0.15cm}
\caption{3D surface map of $\Lambda_1 \Lambda_2$} \label{fig: region comparison and convex cost}
\vspace{-0.5cm}
\end{figure}

\subsubsection{No-switching condition 1)}
If the inequality  $4m_tk_p \leq k_d^2$ holds, the other conditions are rearranged as follows:
\begin{equation} \label{eqn: linear constraints of the first ISS condition}
    \begin{split}
        -b_e (k_f + 1) + k_{d} <  b_f \\
        \Big(\tfrac{k_e}{C} - b_e\Big)k_f + \tfrac{k_e-k_p}{C} + k_d - b_e < b_f
    \end{split}
\end{equation}
where $C \triangleq \tfrac{2K_1}{B_1 - \sqrt{B_1^2-4K_1}} = \tfrac{2k_p}{k_d - \sqrt{k^2_d - 4m_tk_p}}$.

\subsubsection{No-switching condition 2)}
The first and second inequalities are rearranged as follows:
\begin{equation} 
    \begin{split}
        -b_e(1+k_f) + k_d < b_f \\
        -b_e(1+k_f) + 2\sqrt{m_tk_e(1+k_f)} \leq b_f.
    \end{split}
\end{equation}
Meanwhile, the third equation is rearranged as follows:
\begin{equation} \label{eqn: rearranged inequality of the 2nd ISS's 3rd inequality 1}
    (K_1 + K_2)B_2 - 2B_1K_2 < (K_2 - K_1)\sqrt{B_2^2-4K_2}.
\end{equation}
To arrange this inequality in an explicit form, we first need to determine the sign of the left side of (\ref{eqn: rearranged inequality of the 2nd ISS's 3rd inequality 1}). If the left side is negative, the above inequality holds, otherwise, we obtain additional conditions by squaring the both sides. After a few computations, the additional condition is arranged as follows:
\begin{multline} \label{eqn: rearranged inequality of the 2nd ISS's 3rd inequality}
        C_uk_p + (C_lk_e - b_e)(1+k_f) < b_f \\
        < C_lk_p + (C_uk_e - b_e)(1+k_f)
\end{multline}
where
\begin{equation*}
    \begin{split}
        C_l \triangleq \tfrac{B_1 - \sqrt{B_1^2-4K_1}}{2K_1}, \quad C_u \triangleq \tfrac{B_1 + \sqrt{B_1^2-4K_1}}{2K_1}
    \end{split}
\end{equation*}
when $4m_tk_p \leq k_d^2$ holds.

\subsubsection{No-switching condition 3)}
This condition is rearranged in the form of explicit inequalities as follows:
\begin{multline} \label{eqn: linear constraints of the third ISS condition}
    -b_e(1+k_f) + 2\sqrt{m_tk_e(1+k_f)} \leq b_f \\
    \leq -b_e(1+k_f)+k_d.
\end{multline}

\subsubsection{Finite-switching condition}
Fig. \ref{fig: region comparison and convex cost} shows that $\Lambda_1\Lambda_2$ has a bowl shape w.r.t. $k_f$ and $b_f$. Thus, we can find a globally optimal point of $(k_f, b_f)$ which minimizes $J(k_f,b_f)$ defined in step 3) using convex optimization. However, since we cannot differentiate $\Lambda_1\Lambda_2$ due to the modular expression, we need to utilize gradient-free algorithms. Since the given system gets more stable when $\Lambda_1 \Lambda_2$ gets smaller and $(k_f, b_f)$ gets further from their limits $[k_{f,m}, b_{f,m}]$ and $[k_{f,M}, b_{f,M}]$, we set $k_f$ and $b_f$ to the values which minimize the convex cost function $J(k_f, b_f)$ defined in (\ref{eqn: cost function of the fourth ISS condition}) by using the \textit{pattern search} algorithm.

\section{Experimental Results}
This section reports the experimental validation of the proposed motion/force control strategy.
\subsection{Experimental Setups}
The experimental setup for this research consists of four parts: an underactuated coaxial octocopter, a robotic arm, a 1-axis force sensor and a tilted surface. The coaxial octocopter which weighs 3.78kg was assembled with the custom-built frame, eight KDE2314XF-965 motors with corresponding KDEXF-UAS35 electronic speed controllers, and 9-inch APC LPB09045MR propellers, two Turnigy LiPo batteries for power supplement, and Intel NUC for computing. On Intel NUC, Robot Operating System (ROS) is installed in Ubuntu 18.04, and the position controller for the octocopter, servo-angle controller for the robotic arm and the navigation algorithm with Optitrack are executed. The attitude controllers are executed in Pixhawk 4 which is connected to the Intel NUC. The robotic arm is comprised of ROBOTIS dynamixel XH540 and XM430 servo motors. We mount Honeywell FSS2000NSB 1-axis force sensor to the end-effector which is connected to the arduino nano board. The tilted surface is made of medium density fibreboard (MDF) and we attach four Optitrack markers to that surface to measure its rotation matrix $[B_f \ B_m]$. 

The values of the parameters during the experiments are arranged in Table \ref{table: parameters}.
\begin{table}[h!]
\centering
\begin{tabular}{|c | c | c | c| c| c| c|} 
 \hline
 \multicolumn{7}{|c|}{Estimation of Kelvin-Voigt linear model's parameters} \\
 \hline
 $\mu_1$ & $\mu_2$ & $\rho_M$ & $k_{e,m}$ & $b_{e,m}$ & $k_{e,M}$ & $b_{e,M}$\\ 
 \hline
 0.9996 & 0.9996 & 5000 & 50 & 0.1 & 500 & 1 \\
 \hline
 \multicolumn{7}{|c|}{Reference trajectory generation and controller} \\ [0.5ex] 
 \hline
 $\omega_n$ & $k_p$ & $k_d$ & $K_{m,p}$ & $K_{m,d}$ & &\\
 \hline
 10.0 & 23.5 & 19.5 & 23.5$I_2$ & 19.5$I_2$ & &\\
 \hline
 \multicolumn{7}{|c|}{Force-controller-gain scheduler} \\ [0.5ex] 
 \hline
 $k_{f,m}$ & $b_{f,m}$ &$k_{f,M}$ & $b_{f,M}$ & & &\\
 \hline
 0.1 & 10 & 1 & 40 & & &\\
 \hline
\end{tabular}
\caption{Parameters used in the experiments.}
\label{table: parameters}
\end{table}
\vspace{-0.5cm}

We assumed that only the orientation of the contact surface is known while its position is not given. Also, for the constant reference force, $f_{f,r}(t)$ was set to $-6$ while it was set to $-3.5+2.5\cos{\tfrac{2\pi t}{5}}$ for the time-varying reference force.

\begin{figure*}[t]
\centering
\vspace{0.2cm}
\includegraphics[width = 0.95\textwidth]{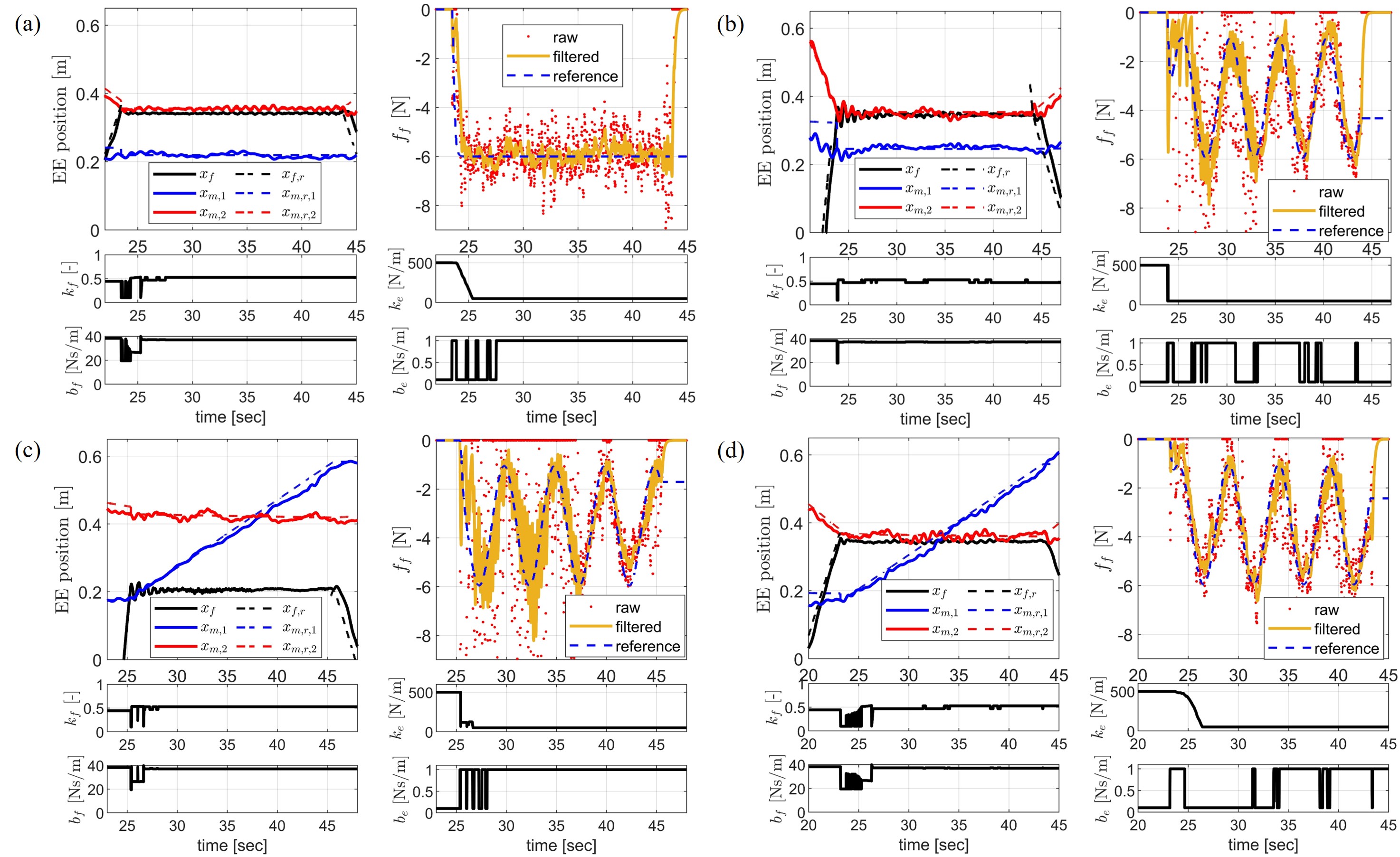}
\vspace{-0.30cm}
\caption{Histories of the end-effector's position ($p_e$), force exerted on the end-effector ($f_f$), force-controller gains ($k_f$ and $b_f$) and estimated environmental parameters ($\hat{k}_e$ and $\hat{b}_e$) of (a) the first one of \textit{Experiment 1}, (b) the second one of \textit{Experiment 1}, (c) the first one of \textit{Experiment 2} and (d) the second one of \textit{Experiment 2}.} \label{fig: experiment_1_and_2}
\vspace{-0.6cm}
\end{figure*}

\subsection{Experiment 1: Force Tracking on the Tilted Surface with Two Different Approach Speeds}
The uUAM configured with an underactuated coaxial octocopter and a robotic arm approaches the tilted surface with two different approach speeds (0.1 m/s and 0.3 m/s) and exerts the constant or time-varying force onto a specific point of that surface.

Fig. \ref{fig: experiment_1_and_2}a presents the measured values of position and exerted force of the end-effector, the force-controller gains and the estimated environment parameters when the uUAM attempts to exert the constant force to the tilted surface after approaching with the slow speed. As observed in the measured force values, the uUAM was able to track the constant reference force trajectory while keeping the constant position of the end-effector.

In Fig. \ref{fig: experiment_1_and_2}b, contrary to Fig. \ref{fig: experiment_1_and_2}a, the reference force trajectory varied with time and the approach speed was relatively fast. Despite the large oscillations in the force measurement, the stable tracking of the time-varying reference force trajectory was finally attained. From this result, we can notice that the proposed controller can also make the uUAM follow the time-varying reference force after the collision with relatively high speed.

Meanwhile, we can also confirm that the proposed force-controller-gain scheduler successfully generated $k_f$ and $b_f$ along with the estimated values of $k_e$ and $b_e$ both in Figs. \ref{fig: experiment_1_and_2}a and \ref{fig: experiment_1_and_2}b. 

\subsection{Experiment 2: Force Tracking while Sliding on the Vertical and Tilted Surfaces}
\subsubsection{Scenario}
In this experiment, the uUAM slides on a vertical or tilted surface while exerting the time-varying force for 20 seconds. The approach speed to the vertical surface is set to 0.3 m/s while that to the tilted surface is set to 0.1 m/s.

\subsubsection{Results}

The result of tracking the time-varying reference force trajectory while sliding on the vertical surface is shown in Fig. \ref{fig: experiment_1_and_2}c. The result of $x_{m,1}$ shows that the uUAM successfully slid in the $+y$ direction of the vertical surface while $f_f$ followed $f_{f,r}$ after the initial oscillation. This result demonstrates that the proposed controller can make the uUAM simultaneously track the time-varying reference motion and force trajectories even with the high approach speed.

Fig. \ref{fig: experiment_1_and_2}d presents the result of tracking the time-varying force while sliding on the tilted surface. The force tracking performance is enhanced than Fig. \ref{fig: experiment_1_and_2}c due to the lower speed.

\section{Conclusions}
This paper presents motion/force control that guarantees a stable contact for an aerial manipulator on an arbitrarily tilted surface. To analyze the dynamic characteristics, the translational dynamic equation w.r.t. the position of the end-effector is derived, and decomposed into force and motion spaces where the force exerted on the end-effector is modeled as the Kelvin-Voigt linear model. Then, we estimate the parameters of Kelvin-Voigt linear model by recursive least-squares estimation, and generate the reference motion and force trajectories based on their setpoints. The disturbance-observer-based controller with scheduling of the force-controller gains is designed based on the stability conditions considering both model uncertainty and switching behavior between the free and contact motion. To check the performance of our controller, we conduct four different force tracking experiments with different approach speeds and reference motion/force trajectories. The results confirm that the proposed controller enables the aerial manipulator to simultaneously track the time-varying reference motion and force trajectories while maintaining stable contact. Future works may involve the design of a switching rule which can enhance the stability during the switch between the free and contact motion or a motion/force control law to push a movable structure.

\normalem
\bibliographystyle{IEEEtran}
\bibliography{myreference}

\end{document}